\documentclass{llncs}

\usepackage[T1]{fontenc}
\usepackage{graphicx}
\usepackage{amssymb}
\usepackage{amsmath}
\usepackage{stmaryrd}
\usepackage[frozencache=true,cachedir=minted-cache]{minted}
\usepackage{url}

\usepackage{enumitem}
\usepackage[colorinlistoftodos, textwidth=3cm]{todonotes} 

\usepackage{cleveref}

\usepackage{tikz}
\usetikzlibrary{decorations.pathmorphing,arrows.meta}
\tikzstyle{node}=[fill=white, draw=black, shape=rectangle, minimum size = 0.4cm, text width=40mm,]
\tikzstyle{circ}=[fill=white, draw=black, shape=ellipse, minimum size = 0.4cm]
\tikzstyle{special}=[draw=black]
\tikzstyle{danger}=[draw=black, dashed]
\pgfdeclarelayer{bg}    % declare background layer
\pgfsetlayers{bg,main}  % set the order of the layers (main is the standard layer)

\usepackage{mathtools}
\usepackage{bbm}
\usepackage{bm}
\usepackage{nicefrac}
\usepackage{thmtools}
\usepackage{thm-restate} % restate theorems with the same number in the appendix

\newcommand{\splus}{\ensuremath{\bm{+}}}
\newcommand{\stimes}{\ensuremath{*}}

\newcommand{\btimes}{\ensuremath{\textstyle\Pi}}

\newcommand{\srzero}{\ensuremath{e_{\oplus}}}
\newcommand{\srone}{\ensuremath{e_{\otimes}}}
\newcommand{\srsplus}{\ensuremath{{\oplus}}}
\newcommand{\srstimes}{\ensuremath{{\otimes}}}
\newcommand{\srbplus}{\ensuremath{{\textstyle\bigoplus}}}

\setitemize{label=--,leftmargin=*}
\setenumerate{leftmargin=*}
\newcommand{\rem}[1]{{\rm #1}}

\def\CKB{\mathfrak{K}}
\newcommand{\Cl}{\mathfrak{C}}
\newcommand{\stru}[1]{\langle #1 \rangle}
\newcommand{\KB}{\mathrm{K}} %knowledge base of module/context
\newcommand{\N}{\boldsymbol{\mathsf{N}}}
\newcommand{\ml}[1]{\mathsf{#1}} %meta-language object
\newcommand{\mi}[1]{\mathit{#1}}
\newcommand{\mlc}{\ml{c}}
\def\isa{\sqsubseteq}
\newcommand{\default}{{\mathrm D}}
\newcommand{\non}{\neg}

\newcommand{\NI}{\mathrm{NI}}
\newcommand{\NR}{\mathrm{NR}}
\newcommand{\NC}{\mathrm{NC}}

\newcommand{\Lcal}{{\cal L}}
\newcommand{\Rcal}{{\cal R}}

\def\Lcals{\Lcal_{\Sigma}}
\def\Lcalsn{\Lcal_{\Sigma,\N}}

\newcommand{\eval}{\textsl{eval}}

\newcommand{\IC}{\mathfrak{I}}
\def\I{\mathcal{I}}
\def\Ic{\I(\mlc)}
\def\Icp{\I(\mlc')}

\newcommand{\vc}[1]{\mathbf{#1}}

\newcommand{\ee}{{\vc{e}}}
\newcommand{\ff}{{\vc{f}}}

\newcommand{\CAS}{\mathit{CAS}}

\def\ICAS{\IC_{\CAS}}

\newcommand{\casmap}{\chi}
\newcommand{\ov}[1]{\overline{#1}}

\title{Contextual Reasoning for Scene Generation}
\subtitle{Technical Report}
\titlerunning{Contextual Reasoning for Scene Generation}

\author{
   Loris~Bozzato\inst{1} \and 
   Thomas~Eiter\inst{2} \and
   Rafael~Kiesel\inst{2} \and 
   Daria~Stepanova\inst{3} 
}

\institute{
  Fondazione Bruno Kessler, 
  Via Sommarive 18, 38123 Trento, Italy \\
  \and
  Institute of Logic and Computation, Technische Universit\"{a}t Wien,\\
  Favoritenstra\ss e 9-11, A-1040 Vienna, Austria\\
  \and
  Bosch Center for Artificial Intelligence, Renningen, Germany
}
\authorrunning{L. Bozzato, T. Eiter, R. Kiesel, D. Stepanova, L. Serafini}

\begin{document}

\maketitle

 \begin{abstract}
   We present a continuation to our previous work, in which we developed the MR-CKR framework to reason with knowledge overriding across contexts organized in multi-relational hierarchies. Reasoning is realized via ASP with algebraic measures, allowing for flexible definitions of preferences.
In this paper, we show how to apply our theoretical work to real autonomous-vehicle scene data. 
%from an external partner (Bosch Deutschland). 
Goal of this work is to apply MR-CKR to the problem of generating challenging scenes for autonomous vehicle learning. In practice, most of the scene data for AV learning models common situations, thus it might be difficult to capture cases where a particular situation occurs (e.g. partial occlusions of a crossing pedestrian). The MR-CKR model allows for data organization exploiting the multi-dimensionality of such data (e.g., temporal and spatial). Reasoning over multiple contexts enables the verification and configuration of scenes, using the combination of different scene ontologies.
% We aim to verify the applicability of our methods to the needs of real datasets by providing a symbolic approach to guide the generation of subsymbolic data for unexpected cases. This will allow us to adapt the definitions for knowledge propagation and preferences and study possibilities for epistemic reasoning.
We describe a framework for
semantically guided data generation,
based on a combination of MR-CKR and Algebraic Measures. The framework is implemented in 
a proof-of-concept prototype exemplifying
some cases of scene generation.

 % \keywords{First keyword  \and Second keyword \and Another keyword.}
 \end{abstract}
%---------------------------------------------------
%---------------------------------------------------
\section{Introduction and motivation}

% # Problem #
Testing and evaluation are important steps in the development and deployment of Automated Vehicles (AVs). 
To comprehensively evaluate the performance of AVs, it is crucial to test the AVs' perception systems in safety-critical scenarios, which rarely happen in naturalistic driving environment, but still possible in practice. 
Therefore, the targeted and systematic generation of such corner cases becomes an
important problem. Most existing studies focus on generating adversarial examples for perception
systems of AVs which are concerned with very simple perturbations in the input (e.g., changing the color or position of a vehicle), whereas limited efforts have been put on the generation of ontology-based and context-specific complex scenes (e.g., child walking a dog in the evening in a rainy weather). 
This is exactly the problem we want to consider in this micro-project.
%is the lack of scene examples for AV learning in the case of uncommon (and dangerous) situations. 
%: in practice, most of the scene data collected for AV learning models common situations, thus it is difficult to have examples of cases where a particular situation occurs.

% # Task definition #
Specifically, we define our task of interest as follows: given an existing scene (represented by a scene graph) from a known dataset, we want to generate a new set of scenes that are variations of the current scene and are:
\begin{enumerate}
\item 
    \textbf{Realistic:} that is, consistent with the ontologies describing objects in the scene (e.g., traffic signs usually do not move);
\item
    \textbf{Interesting:} that is, they satisfy a semantic restriction, which tells us that the scene is for example ``dangerous'' or challenging for our prediction model (e.g., seeing a cat in the middle of the street requires special action);
\item 
    \textbf{Similar:} that is, changing the original scene to the generated scenes requires only small variations.
\end{enumerate}
%
% # Proposed solution #
We propose to use symbolic methods to generate valid and challenging scenes on the base of existing scene graphs and 
semantic definitions of scenes.
In particular, MR-CKR~\cite{BozzatoES:18,DBLP:journals/tplp/BozzatoEK21} is a useful formalism for this. Here, ontological knowledge is contextualized such that in different contexts it may have different interpretations (possibly with non-monotonic effects). This means that MR-CKR can help us in generating \emph{realistic} scenes, since it is capable of handling the background ontologies describing the AC domain. Additionally, we may have different contextualized notions of \emph{interestingness}. MR-CKR also allows us to express this by associating different independent semantic restrictions on scenes within different contexts.

Another benefit of MR-CKR is that it comes with a translation to Answer Set Programming (ASP), which is a declarative programming language that can be used to easily express and efficiently solve hard logical problems. 

Apart from realism and interest, we care about similarity. Thus, we need a way to measure how similar the generated scenes are to the original scene that we started from. So-called Algebraic Measures~\cite{EiterK:20} are of great use here. They are a general framework from the field of ASP that allows us to measure quantities associated with solutions. As such, they are also capable of expressing a similarity measure of scenes, which we need. 

Our main contributions are as follows:
\begin{itemize}
    \item We provide a novel framework for semantically guided data generation, which can be adversarial or training data.
    \item By basing our framework on a combination of MR-CKRs and Algebraic Measures we obtain a highly flexible approach with efficient solving options by employing translations to ASP.
    \item MR-CKRs allow us (i) to incorporate ontological background knowledge ensuring realism of the generated data and (ii) to contextualize the notion of what makes a generated input interesting.
    \item Algebraic Measures enable the maximization of similarity between original and generated data.
    \item Our prototype for scene generation in the domain of AV is intentionally kept minimal but shows promise.
\end{itemize}

\paragraph{Related Work.}
The generation of adversarial or challenging examples for neural models is an important problem that gained interest both in industry\footnote{\url{https://www.efemarai.com/}} and research~\cite{rozsa2016adversarial,yang2023adversarial,chen2023adversarial,gao2018blackbox}.
Also in these works the generation of inputs that are similar to the original ones and realistic is of importance.
However, instead of using symbolic methods to generate new inputs and to verify that the generated inputs are realistic, numerical methods are used here. E.g., \cite{rozsa2016adversarial} uses small numerical perturbations of images, \cite{yang2023adversarial} uses an optimization that minimizes the numerical change of the input data such that it leads to a different prediction of the network. The closest work that we found to ours is \cite{gao2018blackbox}, which generates adversarial text for natural language processing by performing minimal replacements of characters. However, while this optimization for the minimal replacement can be seen as a symbolic approach, no verification of how realistic the newly generated text is, was performed.
% We can consider to limit the problem to a specific case of "dangerous" scenes (by providing a measure or features definition of preferred output scenes).

% We could consider scenes as single images or multiple images with spatial and temporal information and consider "dangerous situations" as "dangerous actions" performed by some actors/elements of the scene.
\section{Framework Overview}
\begin{figure}[tbp]%[thbp]
    \centering
    % Link to modify 
    % https://www.mathcha.io/editor/30l5KHeMi1zCpmu41961XH8rqVZdUOmqNgwiyOVXX
    \tikzset{every picture/.style={line width=0.75pt}} %set default line width to 0.75pt        
    
    \begin{tikzpicture}[x=0.75pt,y=0.75pt,yscale=-1,xscale=1]
    %uncomment if require: \path (0,328); %set diagram left start at 0, and has height of 328
    
    %Rounded Rect [id:dp43630665288160597] 
    \draw   (57.17,85.13) .. controls (57.17,66.47) and (72.3,51.33) .. (90.97,51.33) -- (266.87,51.33) .. controls (285.53,51.33) and (300.67,66.47) .. (300.67,85.13) -- (300.67,186.53) .. controls (300.67,205.2) and (285.53,220.33) .. (266.87,220.33) -- (90.97,220.33) .. controls (72.3,220.33) and (57.17,205.2) .. (57.17,186.53) -- cycle ;
    %Rounded Rect [id:dp6676670970986465] 
    \draw   (135,163) .. controls (135,159.13) and (138.13,156) .. (142,156) -- (183,156) .. controls (186.87,156) and (190,159.13) .. (190,163) -- (190,184) .. controls (190,187.87) and (186.87,191) .. (183,191) -- (142,191) .. controls (138.13,191) and (135,187.87) .. (135,184) -- cycle ;
    %Rounded Rect [id:dp6103371846306743] 
    \draw   (239,164) .. controls (239,160.13) and (242.13,157) .. (246,157) -- (287,157) .. controls (290.87,157) and (294,160.13) .. (294,164) -- (294,185) .. controls (294,188.87) and (290.87,192) .. (287,192) -- (246,192) .. controls (242.13,192) and (239,188.87) .. (239,185) -- cycle ;
    %Rounded Rect [id:dp6897244260566041] 
    \draw   (43.67,70.93) .. controls (43.67,47.96) and (62.29,29.33) .. (85.27,29.33) -- (489.07,29.33) .. controls (512.04,29.33) and (530.67,47.96) .. (530.67,70.93) -- (530.67,195.73) .. controls (530.67,218.71) and (512.04,237.33) .. (489.07,237.33) -- (85.27,237.33) .. controls (62.29,237.33) and (43.67,218.71) .. (43.67,195.73) -- cycle ;
    %Rounded Rect [id:dp9482591790763364] 
    \draw   (306,84.53) .. controls (306,65.65) and (321.31,50.33) .. (340.2,50.33) -- (482.47,50.33) .. controls (501.35,50.33) and (516.67,65.65) .. (516.67,84.53) -- (516.67,187.13) .. controls (516.67,206.02) and (501.35,221.33) .. (482.47,221.33) -- (340.2,221.33) .. controls (321.31,221.33) and (306,206.02) .. (306,187.13) -- cycle ;
    %Rounded Rect [id:dp23126006571973756] 
    \draw   (71.17,106.8) .. controls (71.17,103.04) and (74.21,100) .. (77.97,100) -- (277.2,100) .. controls (280.96,100) and (284,103.04) .. (284,106.8) -- (284,127.2) .. controls (284,130.96) and (280.96,134) .. (277.2,134) -- (77.97,134) .. controls (74.21,134) and (71.17,130.96) .. (71.17,127.2) -- cycle ;
    %Rounded Rect [id:dp05203456975997234] 
    \draw   (329.11,165.25) .. controls (329.11,160.18) and (333.22,156.08) .. (338.28,156.08) -- (488.83,156.08) .. controls (493.89,156.08) and (498,160.18) .. (498,165.25) -- (498,192.75) .. controls (498,197.82) and (493.89,201.92) .. (488.83,201.92) -- (338.28,201.92) .. controls (333.22,201.92) and (329.11,197.82) .. (329.11,192.75) -- cycle ;
    %Straight Lines [id:da5287685410253293] 
    \draw    (135,171) -- (102.68,191.91) ;
    \draw [shift={(101,193)}, rotate = 327.09] [color={rgb, 255:red, 0; green, 0; blue, 0 }  ][line width=0.75]    (10.93,-3.29) .. controls (6.95,-1.4) and (3.31,-0.3) .. (0,0) .. controls (3.31,0.3) and (6.95,1.4) .. (10.93,3.29)   ;
    %Straight Lines [id:da5029866295366923] 
    \draw    (240,174) -- (207.68,194.91) ;
    \draw [shift={(206,196)}, rotate = 327.09] [color={rgb, 255:red, 0; green, 0; blue, 0 }  ][line width=0.75]    (10.93,-3.29) .. controls (6.95,-1.4) and (3.31,-0.3) .. (0,0) .. controls (3.31,0.3) and (6.95,1.4) .. (10.93,3.29)   ;
    %Curve Lines [id:da3834821184244326] 
    \draw    (190,163) .. controls (230,133) and (199,194) .. (239,164) ;
    %Rounded Rect [id:dp9879804479740122] 
    \draw   (329.11,101.94) .. controls (329.11,97.06) and (333.07,93.1) .. (337.95,93.1) -- (489.16,93.1) .. controls (494.04,93.1) and (498,97.06) .. (498,101.94) -- (498,128.46) .. controls (498,133.34) and (494.04,137.3) .. (489.16,137.3) -- (337.95,137.3) .. controls (333.07,137.3) and (329.11,133.34) .. (329.11,128.46) -- cycle ;
    %Straight Lines [id:da818997066160482] 
    \draw  [dash pattern={on 0.84pt off 2.51pt}]  (164.17,134.5) -- (160.56,152.54) ;
    \draw [shift={(160.17,154.5)}, rotate = 281.31] [color={rgb, 255:red, 0; green, 0; blue, 0 }  ][line width=0.75]    (10.93,-3.29) .. controls (6.95,-1.4) and (3.31,-0.3) .. (0,0) .. controls (3.31,0.3) and (6.95,1.4) .. (10.93,3.29)   ;
    %Straight Lines [id:da12512902102817336] 
    \draw  [dash pattern={on 0.84pt off 2.51pt}]  (226.17,134.5) -- (263.42,155.52) ;
    \draw [shift={(265.17,156.5)}, rotate = 209.43] [color={rgb, 255:red, 0; green, 0; blue, 0 }  ][line width=0.75]    (10.93,-3.29) .. controls (6.95,-1.4) and (3.31,-0.3) .. (0,0) .. controls (3.31,0.3) and (6.95,1.4) .. (10.93,3.29)   ;
    %Rounded Rect [id:dp29516364927344985] 
    \draw   (165.17,70.1) .. controls (165.17,67.28) and (167.45,65) .. (170.27,65) -- (268.07,65) .. controls (270.88,65) and (273.17,67.28) .. (273.17,70.1) -- (273.17,85.4) .. controls (273.17,88.22) and (270.88,90.5) .. (268.07,90.5) -- (170.27,90.5) .. controls (167.45,90.5) and (165.17,88.22) .. (165.17,85.4) -- cycle ;
    
    % Text Node
    \draw (93,57) node [anchor=north west][inner sep=0.75pt]   [align=left] {MR-CKR };
    % Text Node
    \draw (150,162) node [anchor=north west][inner sep=0.75pt]   [align=left] {c1};
    % Text Node
    \draw (254,163) node [anchor=north west][inner sep=0.75pt]   [align=left] {c2};
    % Text Node
    \draw (288,34) node [anchor=north west][inner sep=0.75pt]   [align=left] {ASP};
    % Text Node
    \draw (313.44,59.65) node [anchor=north west][inner sep=0.75pt]   [align=left] {Additional ASP Constraints};
    % Text Node
    \draw (81,105.8) node [anchor=north west][inner sep=0.75pt]   [align=left] {Possible Scene Modification};
    % Text Node
    \draw (343.89,156.45) node [anchor=north west][inner sep=0.75pt]   [align=left] {\begin{minipage}[lt]{90.07pt}\setlength\topsep{0pt}
    \begin{center}
    Strong Constraints:\\Danger Presence
    \end{center}
    
    \end{minipage}};
    % Text Node
    \draw (62,191) node [anchor=north west][inner sep=0.75pt]   [align=left] {Danger 1};
    % Text Node
    \draw (167,194) node [anchor=north west][inner sep=0.75pt]   [align=left] {Danger 2};
    % Text Node
    \draw (343.33,93.48) node [anchor=north west][inner sep=0.75pt]   [align=left] {\begin{minipage}[lt]{96.85pt}\setlength\topsep{0pt}
    \begin{center}
    Weak Constraints:\\Similarity Preference
    \end{center}
    
    \end{minipage}};
    % Text Node
    \draw (168.27,66) node [anchor=north west][inner sep=0.75pt]   [align=left] {Base Ontology};

    \end{tikzpicture}

    \caption{The general framework for generating (similar) dangerous scenes with ASP according to (possibly) related types of dangers defined by an MR-CKR.}
    \label{fig:framework}
\end{figure}
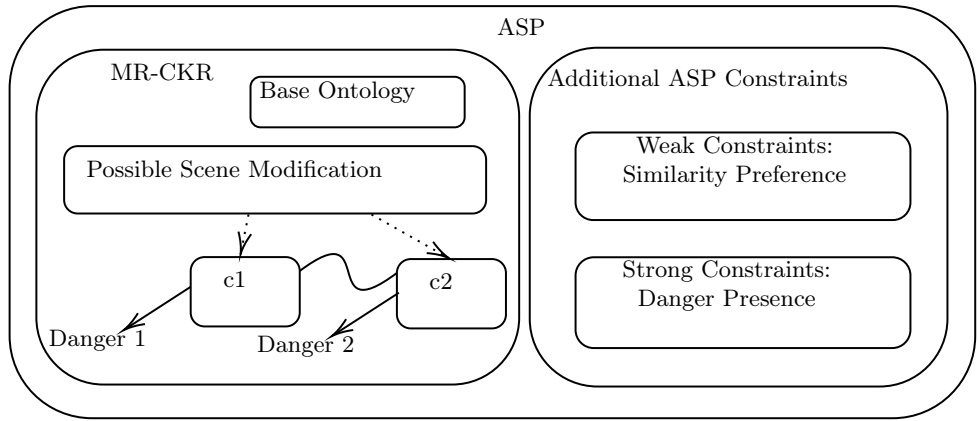
Before we go into the technical details of how we generate descriptions of new challenging training scenes, we provide a general structural overview of our framework.

We consider for this, the schema described in \Cref{fig:framework}. Here, we see that we use an MR-CKR to define, on the one hand, the possible scene modifications and on the other hand different contexts, here $c1$ and $c2$, that specify possibilities for a scene to be dangerous/interesting. Optimally, these different types of danger correspond to diagnoses of a neural network engineer for poor performance of the current neural network. For example, in the AV context, we might observe that a car does not stop in the correct location when there is not only a stop sign but also a stop line marking that specifies where the car should stop. Here, we would therefore want to modify scenes in such a manner that they have both a stop sign and a stop line marking.

Generally, the goal is to generate more scenes that we suspect the network also performs badly on, such that we have adversarial examples that we can use to train the neural network in the hopes of improving its performance on these situations that are hard for it, due to a lack of training data. Given the definitions of danger in different contexts (i.e., based on different diagnosis) that may be related via specialization or otherwise, we can then obtain an equivalent encoding in ASP to obtain models, i.e., generated scenes that are realistic according to the base ontology included in the MR-CKR. 

Additionally, we add further ASP constraints to make sure the modifications of the scene are such that the resulting scene is (a) dangerous (using the strong constraints) and (b) as similar as possible to a given starting scene (using the weak constraints that express the algebraic measure).

Putting both things together, we can thus obtain realistic, dangerous scenes that are as similar to the starting scene as possible. On top of that, the different contexts allow us to specify different types of target dangers resulting in one generated scene that includes it per context.

In the following, we substantiate our abstract idea by formalizing how we generate scenes with MR-CKR and measure their similarity with Algebraic Measures.
%------------------------------------------
\section{Formalization of scene generation problem in MR-CKR}
We begin by introducing formally the MR-CKR framework 
and we provide a solution making use of MR-CKR in scene generation.

\subsection{MR-CKR definition}
We assume the customary definitions for description logics (see, e.g.,~\cite{dlhb} for an introduction).
We summarize in the following the main definitions of MR-CKR 
(as introduced in~\cite{DBLP:journals/tplp/BozzatoEK21}). 
% and provide a formalization of the scene generation problem in terms of this framework.

% # Contextual relation #
We consider a generic description language $\Lcals$ based
on a DL signature $\Sigma$, which is composed of
a set of concept names $\NC$, role names $\NR$
and individual names $\NI$.

Consider a nonempty set $\N \subseteq \NI$  of \emph{context names}.
A \emph{contextual relation} is any strict order $\prec_i \subseteq \N \times \N$ over contexts.
A way to define contextual relations 
is to use \emph{contextual dimensions}~\cite{DBLP:conf/kr/BozzatoSE18,serafini-homola-ckr-jws-2012}, that is a set of contextual ``coordinates'' associated
to each of the contexts: in the case of scene descriptions,
for example, these can represent the time of the day, location type or situation occurring in a scene. 
The contextual structure, then, is defined 
from the product of order of features (dimensions) associated to  the contexts, corresponding to a contextual relation.

In a MR-CKR, axioms inside contexts can be specified as
defeasible (i.e. they can be overridden in case of exceptions)
with respect to one of the contextual relations composing the
contextual structure.

\begin{definition}[r-defeasible axiom]
Given a set $\Rcal$ of contextual relations over $\N$ and a description language $\Lcals$,
an \emph{r-defeasible axiom} is any expression of the form
$\default_r(\alpha)$, where $\alpha$ is an axiom of $\Lcals$
and $\prec_r \in \Rcal$.
\end{definition}
We allow for the use of r-defeasible axioms in the local language
of contexts:

\begin{definition}[contextual language]
Given a set of context names $\N$,
for every description language $\Lcals$
we define $\Lcalsn$ as the extension of
$\Lcals$ where:
(i) $\Lcalsn$ contains the set of r-defeasible axioms in $\Lcals$;
(ii) $\eval(X,\mlc)$ is a concept (resp.\ role) of $\Lcalsn$ if
$X$ is a concept (resp.\ role) of $\Lcals$ and $\mlc \in \N$. 
\end{definition}
Multi-relational CKRs are then composed by a global structure
of context based on the contextual relations in $\Rcal$ and
a set of DL knowledge bases associated to each of the local
contexts.

\begin{definition}[multi-relational simple CKR]
A \emph{multi-relational simple CKR (sCKR)} over
$\Sigma$ and $\N$ is a structure $\CKB = \stru{\Cl, \KB_{\N}}$ where:
\begin{itemize}
  \item 
     $\Cl$ is a structure $(\N, \prec_1, \dots, \prec_m)$
		 where each $\prec_i$ is a contextual relation over
                 $\N$, and
  \item
    $\KB_{\N} = \{\KB_\mlc\}_{\mlc \in \N}$ for each context name $\mlc \in \N$,
    $\KB_\mlc$ is a DL knowledge base over $\Lcalsn$. %$\Lcal^e_\Sigma \cup \Lcal_\Sigma^\default$
  \end{itemize}  
\end{definition}
% A sCKR that combines temporal and coverage orderings can be defined by %a global context 
% $\Cl = (\N, \prec_t, \prec_c)$.
% %
% For simplicity, we assume that the priority for the combination of orderings is defined by the 
% linear order in which they appear in %the definition of 
% $\Cl$: 
% in the case above, we prioritize %the temporal relation 
% $\prec_t$ over %the coverage 
% $\prec_c$. 

\begin{example}
   We provide a simple example of MR-CKR
   to better explain the intended use of 
   defeasible axioms. 
   Consider the sCKR $\CKB = \stru{\Cl, \{\KB_1, \KB_2\}}$ composed by the following elements:
   \[
   \begin{array}{rl}
        \Cl = &  \{ \mlc_2 \prec_c \mlc_1\} \\
        \KB_1 = & \{ \default_c(\mi{Dog} \isa \non \mi{DangerousAnimal} )\}\\
        \KB_2 = & \{ \mi{Dog} \isa \mi{DangerousAnimal}, \mi{Dog(d)}\}\\
   \end{array}
   \]
   Intuitively, we want to recognize that in the more specific context $\mlc_2$, dogs are considered as
   dangerous animals, thus the more general
   defeasible axiom in $\mlc_1$ is not applied to 
   the instance $d$ of $\mi{Dog}$.
\end{example}

Interpretations of MR-CKRs are family of DL interpretations
associated to each of the contexts.

\begin{definition}[sCKR interpretation]
\label{def:ckr-int}
  An interpretation for $\Lcalsn$ is a family $\IC = \{\Ic\}_{\mlc\in\N}$ of $\Lcals$ interpretations, such that
  $\Delta^{\Ic}\,{=}\, \Delta^{\Icp}$ and $a^{\Ic} \,{=}\, a^{\Icp}$, for
    every $a \,{\in}\, \NI$ and $\mlc,\mlc' \,{\in}\, \N$.
\end{definition}

The interpretation of concepts and role expressions in $\Lcalsn$
is obtained by extending the standard interpretation 
to $\eval$ expressions:
for every $\mlc \in \N$, $\eval(X,\mlc')^{\Ic} = X^{\Icp}$.
%
% # CAS interpretation #
We consider the definition of axiom instantiation 
provided by~\cite{BozzatoES:18}:
given an axiom $\alpha \in \Lcal_\Sigma$ with FO-translation
$\forall\vc{x}.\phi_\alpha(\vc{x})$, the \emph{instantiation} of $\alpha$
with a tuple $\ee$ of individuals in $\NI$, written $\alpha(\ee)$, is the
specialization of $\alpha$ to $\ee$, i.e., $\phi_\alpha(\ee)$,
depending on the type of $\alpha$.

A \emph{clashing assumption} for a context $\mlc$ and contextual relation $r$
is a pair $\stru{\alpha, \ee}$
such that $\alpha(\ee)$ is an axiom instantiation of $\alpha$, and
\mbox{$\mlc' \succeq_{-r} \mlc'' \succ_r \mlc$.}
A \emph{clashing set} for 
$\stru{\alpha,\ee}$
is a satisfiable set $S$ of ABox assertions s.t.
 $S \cup \{\alpha(\ee)\}$ is unsatisfiable.

\begin{definition}[CAS-interpretation]
  A \emph{CAS-interpretation} is a structure\linebreak 
	$\ICAS=\stru{\IC,\ov{\casmap}}$ where
  $\IC$ is an interpretation and
	$\ov{\casmap} = \{\casmap_1, \dots, \casmap_m\}$ such that each 
	$\casmap_i$, for $i \in \{1, \dots, m\}$, maps every $\mlc \in \N$
  to a set $\casmap_i(\mlc)$ of clashing assumptions for context $\mlc$
	and context relation $\prec_i$. 
\end{definition}

\begin{definition}[CAS-model]
\label{def:cas-model}
Given a multi-relation sCKR $\CKB$,
a CAS-interpretation %\linebreak 
$\ICAS = \stru{\IC, \ov{\casmap}}$
is a \emph{CAS-model} for $\CKB$ (denoted $\ICAS \models \CKB$), 
if the following holds:
% \footnote{Here, it is important to ensure that (defeasible) axioms are correctly propagated w.r.t.\ \emph{any} context relation $\prec_{i}$.}: 
\begin{enumerate}[label=(\roman*)]
 \item
   for every $\alpha \in \KB_\mlc$ (strict axiom), 
   and $\mlc' \preceq_{*}\mlc$, $\Icp \models \alpha$;
 \item
   for every $\default_i(\alpha) \in \KB_\mlc$ 
	 and $\mlc' \preceq_{-i} \mlc$, $\I(\mlc') \models \alpha$;
  \item
    for every $\default_i(\alpha) \in \KB_\mlc$ 
    and $\mlc'' \prec_i \mlc' \preceq_{-i} \mlc$,    
		if $\stru{\alpha,\vc{d}} \notin \casmap_i (\mlc'')$,
    then $\I(\mlc'') \models \phi_\alpha(\vc{d})$. 
\end{enumerate}
\end{definition}

We provide a \emph{local preference}\/ on clashing assumption sets for each of the relations:

\begin{list}{\emph{(LP).}}{\setlength{\topsep}{2pt}
\setlength{\leftmargin}{14pt}
\setlength{\itemsep}{0pt}
\setlength{\itemindent}{10pt}}
\item[\emph{(LP).}]  $\chi_i^1(\mlc) > \chi_i^2(\mlc)$, if for every $\stru{\alpha_1,\ee} \in \chi_i^1(\mlc) \setminus \chi_i^2(\mlc)$ with 
  $\default_i(\alpha_1)$ at a context $\mlc_1 \succeq_{-i} \mlc_{1b} \succ_{i}  \mlc$, 
  some $\stru{\alpha_2,\ff} \in \chi_i^2(\mlc) \setminus
  \chi_i^1(\mlc)$ exists with 
	$\default_i(\alpha_2)$ at context $\mlc_2 \succeq_{-i} \mlc_{2b} \succ_i \mlc$ 
	s.t.\ $\mlc_{1b} \succ_i \mlc_{2b}$.
\end{list}
Intuitively, $\chi^1_i(\mlc)$ is preferred to $\chi^2_i(\mlc)$ if 
$\chi^1_i(\mlc)$ exchanges the ``more costly'' exceptions of $\chi^2_i(\mlc)$
at more specialized contexts with ``cheaper'' ones at more general contexts. 
%
% As above, multiple options for 
% local preference can be adopted, 
% cf.~\cite{DBLP:conf/kr/BozzatoSE18}
% for ranked hierachies.

% # Justification #
Two DL interpretations $\I_1$ and $\I_2$ are
\emph{$\NI$-congruent}, if $c^{\I_1} = c^{\I_2}$ holds for every $c\in \NI$.
This extends to CAS interpretations $\IC_{\CAS} = \stru{\IC, \ov{\casmap}}$ by considering all context interpretations $\I(\mlc) \in \IC$.

\begin{definition}[justification]
	We say that $\stru{\alpha, \ee} \in \chi_i(\mlc)$ is
	\emph{justified} for a $\CAS$ model $\IC_{\CAS}$,   
	if some clashing set
	$S_{\stru{\alpha,\ee},\mlc}$ exists
	such that, for every %CAS-model
	$\IC_{\CAS}' = \stru{\IC',\ov{\casmap}}$ of $\CKB$ that is $\NI$-congruent with $\IC_{\CAS}$, 
	it holds that $\I'(\mlc) \models S_{\stru{\alpha,\ee},\mlc}$.
	A $\CAS$ model $\ICAS$ of 
  a sCKR $\CKB$ is \emph{justified}, 
  if every $\stru{\alpha, \ee} \in \ov{\casmap}$ is justified in $\CKB$.
\end{definition}

% # Model preference #
We define a \emph{model preference} 
by combining the preferences of the relations:
it is a global lexicographical ordering on models where
each $\prec_i$ defines the ordering at the $i$-th position.

\smallskip\noindent
\emph{(MP).}\
	$\IC^1_{\CAS} = \stru{\IC^1, \casmap^1_1, \dots, \casmap^1_m}$ 
	is preferred to $\IC^2_{\CAS} = \stru{\IC^2, \casmap^2_1, \dots, \casmap^2_m}$ if
  \begin{enumerate}[label=(\roman*)]
	\item 
	   there exists $i \in \{1, \dots, m\}$ and some $\mlc \in \N$ s.t. 
	   $\casmap^1_i(\mlc) > \casmap^2_i(\mlc)$ and not $\casmap^2_i(\mlc) > \casmap^1_i(\mlc)$, and
     for no context $\mlc' \neq \mlc \in \N$ it holds that 
	   $\casmap^1_i(\mlc') < \casmap^2_i(\mlc')$ and not $\casmap^2_i(\mlc') < \casmap^1_i(\mlc')$.
  \item
	   for every $j < i \in \{1, \dots, m\}$, 
		 it holds $\casmap^1_j \approx \casmap^2_j$
		 (i.e.\ %condition 
		 (i) or its converse 
		 do not hold
		 for $\prec_j$).
\end{enumerate}

% # CKR model #
\begin{definition}[CKR model]
\label{def:ckr-model}
  An interpretation $\IC$ is a \emph{CKR model} of a sCKR $\CKB$ 
	(in symbols, $\IC\models\CKB$) if:
 (i) $\CKB$ has some justified CAS model $\ICAS = \stru{\IC, \ov{\casmap}}$;	  
(ii) there exists no justified %CAS model 
     $\ICAS' = \stru{\IC', \ov{\casmap}'}$
     that is preferred to $\ICAS$.
\end{definition}

\begin{example}
   Using the semantics mechanism shown above, we
   can show how to interpret the sCKR in previous example.
   In particular, we can consider the CAS-interpretation
   $\ICAS = \stru{\IC, \ov{\casmap}}$
   where $\chi(\mlc_2) = \{\stru{\mi{Dog} \isa \non \mi{DangerousAnimal}, d}\}$.
   This implies that $\ICAS$ is a CAS-model if
   the defeasible axiom of $\mlc_1$ is not applied
   to the only $\mi{Dog}$ in $\mlc_2$, as expected.
   Note that such CAS-model is also justified,
   since the clashing assumption 
   admits the clashing set $\{\mi{Dog}(d), \mi{DangerousAnimal}(d)\}$: thus, considering that
   no other alternative CAS-model that is minimal with
   respect to the preference can be defined, the considered
   interpretation is also a CKR-model.
   The preference defined above is useful to
   prefer defeasible axioms in the most specific
   contexts: for example, if $\mi{Dog} \isa \mi{DangerousAnimal}$ in $\mlc_2$ was defined as defeasible (w.r.t. the same contextual relation of the above defeasible axiom), the context below $\mlc_2$
   would have preferred the more specific axiom and thus 
   the interpretations where exceptions are 
   made on the more generic axiom of the upper context
   $\mlc_1$.
\end{example}

As a method to implement reasoning on MR-CKR,
in~\cite{DBLP:journals/tplp/BozzatoEK21}
we provided a translation for MR-CKRs to ASP logic
programs: in particular, we showed that such translation
can be used to reason on instance checking and query answering
in a given context. 

%- - - - - - - - - - - - - - - - - - - - - - - - - - 
\subsection{Scene generation in MR-CKR}

Following the intuitive structure of Figure~\ref{fig:framework},
the role of MR-CKR in our architecture is to define the logical constraints of the scenes we want to generate, on the base of a common scene ontology.
% in our case the scene ontology is~\cite{chowdhury2021towards}

Given its multi-contextual structure, the MR-CKR is useful to
provide a complex representation (a contextualization) of the contents of the base scene and its modifications towards the different diagnoses of interest.

\begin{figure}[thbp]
       \centering
       \includegraphics[trim=165 175 165 135, clip, width=\textwidth]{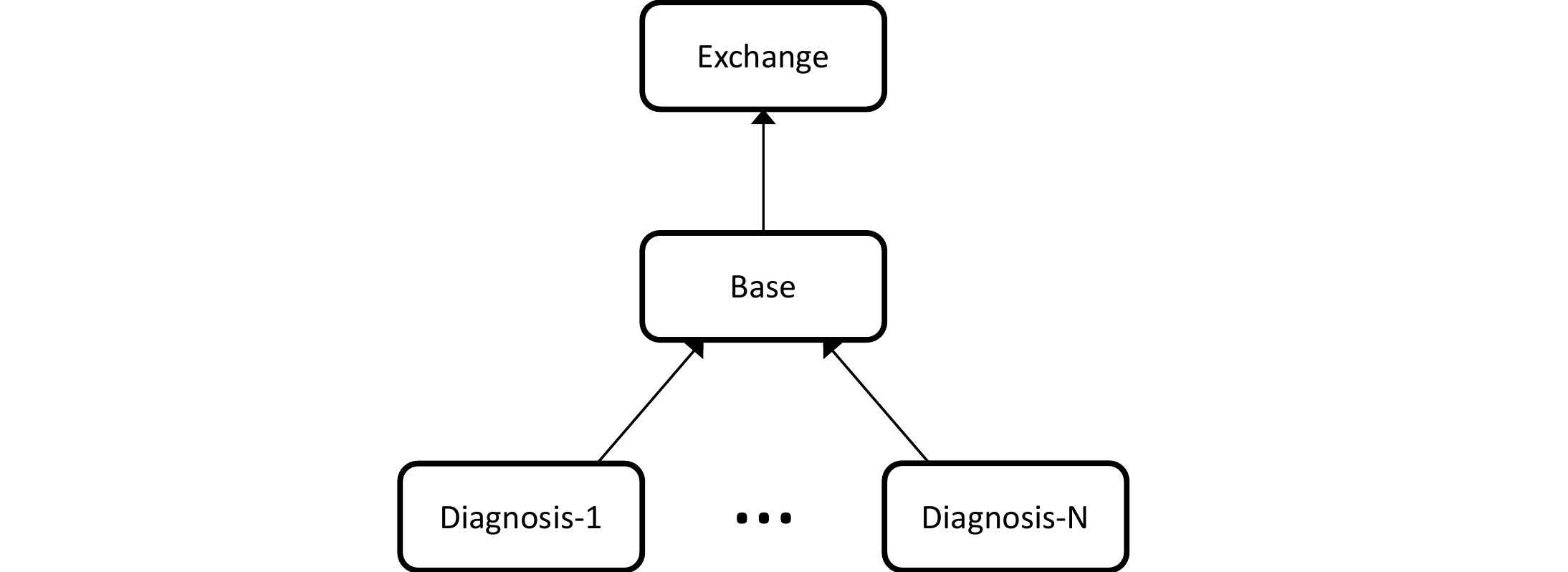}
       \caption{General structure of contexts for scene modification}
       \label{fig:base-mrckr}
   \end{figure}

With respect to the second aspect, the basic organization of contexts can be defined as in Figure~\ref{fig:base-mrckr}.

\noindent
The contexts of this structure are related by a contextual relation $\succ_{sim}$, denoting the relation of similarity: the upper context $\mi{Exchange}$ contains, in form of defeasible axioms, the axioms that can be modified in the diagnosis scenes.
In the $\mi{Base}$ context, we assume to have the description of the base scene and the base axioms of the scene description ontology.
The contexts $\mi{Diagnosis\textit{-}1}, \dots, \mi{Diagnosis\textit{-}N}$, then, 
provide the different modifications to the base scene we are interested to model.
The kind of axioms that are needed to model the different modifications depend on the kind of changes (additions, deletions, etc.) that we want to admit in scene modifications: more detail on such axioms will be provided in the following sections, where we consider specific modifications.

With respect to the scene contextualization, we can make use of the multi-relational nature of MR-CKR to further define the context in which the scene take place.

\begin{example}
An example of such contextualization is shown in Figure~\ref{fig:scene-mrckr}.

   \begin{figure}[thbp]
       \centering
       \includegraphics[trim=0 175 0 65, clip, width=\textwidth]{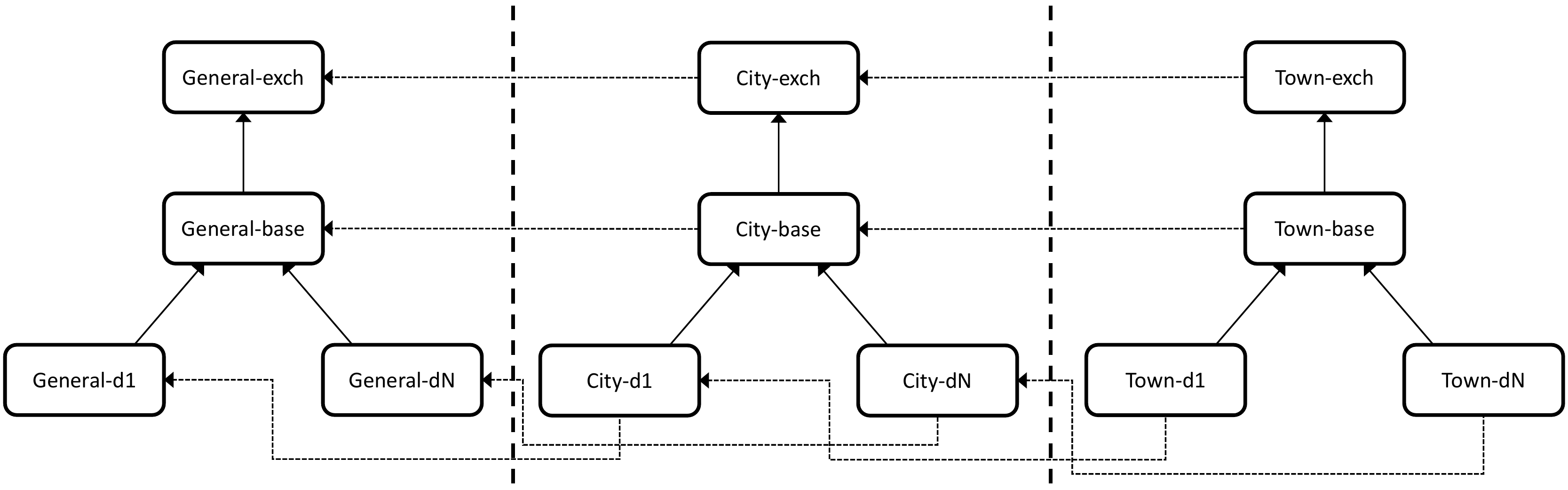}
       \caption{Example of general structure of MR-CKR for scene representation}
       \label{fig:scene-mrckr}
   \end{figure}   

In this contextual structure, the relation represented by the horizontal arrows represent the specialization of scenes with respect to the specificity of the location: starting from axioms that are verified for \emph{general} scenes, we can add further logical constraints that are true for \emph{city} scenes and 
then \emph{town} scenes. Note that such direction is orthogonal to the base contextual structure described above.   
\end{example}
After modelling scenes by such framework, we want to use the
translation of MR-CKR to ASP in order to generate
the possible models of the diagnoses contexts:
these models then correspond to alternative generated scenes.
However, we now need a method to provide a measure the \emph{similarity} of the generated scenes with respect to the scenes of interest: as we detail in the following sections, this can be easily obtained by means of algebraic measures.

% \lbnote{Target context $\mlc \in \N$, scene target description TBox $T$}

% \begin{definition}[Scene generation problem]
%   Given a MR-CKR $\CKB$, a target context 
%   $\mlc \in \N$ and a target scene description $T$ (in the language of $\mlc$).
%   The scene generation problem ask for the set of
%   models $M$ of $\CKB \cup T$ such that
%   $M \models \mlc : \ml{TargetScene}(s)$
% \end{definition}

% \lbnote{We need to fix the vocabulary for the scenes and their relations: $\ml{Scene}, \ml{Object}, \ml{contains}$ ...}

%############################

%------------------------------------------
\section{Formalization of Similarity using Algebraic Measures}
If we want to generate a new scene based on a starting scene, we want to optimize a measure of similarity. For measuring similarity, we can make use of algebraic measures. The intuitive idea behind algebraic measures is that they allow us to measure a quantity associated with an interpretation or a model. In order to allow measuring different quantities in a uniform framework, we use the algebraic structure of \emph{semirings}, which allow for many different forms of computation.

\subsection{Preliminaries}
We introduce algebraic measures and their necessary preliminaries. 
\begin{definition}[Monoid]
A \emph{monoid} $\mathcal{M} = (M, \srstimes, \srone)$ consists of an associative \emph{binary operation} $\srstimes$ on a set $M$ with \emph{neutral element} $\srone$, also called \emph{identity element}. Here, a binary operation on $M$ is a function $\srstimes: M \times M \rightarrow M$ that maps pairs of values from $M$ to a value in $M$. We write the application of such a binary operation $\srstimes$ to a pair $(m_1,m_2)$ of values $m_1, m_2 \in M$ in infix notation $m_1 \srstimes m_2$.

A value $\srone \in M$ is a neutral element for a binary operation $\srstimes$ on $M$ if for all values $m \in M$ it holds that
\[
\srone \srstimes m = m = m \srstimes \srone.
\]

Additionally, a binary operation $\srstimes$ on $M$ is \emph{associative}, if for all $m, m', m'' \in M$ it holds that 
\[
    m\srstimes(m' \srstimes m'') = (m \srstimes m') \srstimes m''.
\]
\end{definition}
Some examples of monoids are
\begin{itemize}
    \item $\textsc{Strings} = (\{0,1\}^{*}, \odot, \varepsilon)$, the set of binary strings with concatenation $\odot$ and empty string $\varepsilon$ is a non-commutative, non-idempotent, and non-invertible monoid,
    \item $\mathcal{P}(A) = (2^{A}, \cup, \emptyset)$, the set of subsets for a set $A$ with union $\cup$ is a commutative, idempotent and non-invertible monoid,
    \item $\mathcal{P}(A) = (2^{A}, \cap, A)$, the set of subsets for a set $A$ with intersection $\cap$ is a commutative, idempotent and non-invertible monoid,
    \item $\mathbb{Z} = (\mathbb{Z}, +, 0)$, the integers with addition $+$ is a commutative, non-idempotent and invertible monoid.
\end{itemize}

Based on monoids, we introduce semirings.
\begin{definition}[Semiring]
A \emph{semiring} $\mathcal{R} = (R, \oplus, \otimes, e_{\oplus}, e_{\otimes})$ is a nonempty set $R$ equipped with two binary operations $\oplus$ and $\otimes$, called addition and multiplication, such that
\begin{itemize}
    \item $(R, \oplus)$ is a commutative monoid with identity element $e_{\oplus}$,
    \item $(R, \otimes)$ is a monoid with identity element $e_{\otimes}$,
    \item multiplication left and right distributes over addition, i.e., for all $r, r', r'' \in R$ it holds that
    \begin{align*}
        r \srstimes (r' \srsplus r'') &= r \srstimes r' \srsplus r \srstimes r''\\
        (r' \srsplus r'') \srstimes r &= r' \srstimes r \srsplus r'' \srstimes r
    \end{align*}
    \item and multiplication by $e_{\oplus}$ annihilates $R$, i.e., for all $r \in R$ it holds that
    \[
        r\otimes e_{\oplus} = e_{\oplus} = e_{\oplus}\otimes r.
    \]
\end{itemize} 
\end{definition}
Some examples of semirings are 
\begin{itemize}
    \item $\mathbb{B} = (\{0, 1\}, \vee, \wedge, 0, 1)$, the Boolean semiring, with disjunction and conjunction as addition and multiplication,
    \item $\mathbb{F} = (\mathbb{F}, +, \cdot, 0, 1)$, for $\mathbb{F} \in \{\mathbb{N}, \mathbb{Z}, \mathbb{Q}, \mathbb{R}\}$ the semiring of the numbers in $\mathbb{F}$ with addition and multiplication,
    \item $\mathcal{P}(A) = (2^{A}, \cup, \cap, \emptyset, A)$, the semiring over the powerset of $A$ with union and intersection, and
    \item $\mathcal{R}_{\min, +} = (\mathbb{N}\cup\{\infty\}, \min, +, \infty, 0)$, the min-plus semiring.
\end{itemize}
Another list of semirings, which is annotated with applications, can be found in~\cite{kimmig2017algebraic}.

In order to connect the quantitative aspects of semirings and the qualitative ones of logics we use weighted logics. They were initially introduced by \cite{droste2007weighted} in the second order setting. Here, we only introduce the restricted version for propositional logic.
\begin{definition}[Syntax]
Let $\mathcal{V}$ be a set of propositional variables and let $\mathcal{R} = (R, \srsplus, \srstimes, \srzero, \srone)$ be a semiring. A \emph{weighted} (propositional) formula over $\mathcal{R}$ is of the form $\alpha$ given by the grammar
\begin{align*}
    \alpha ::= k &\mid v \mid \neg v \mid \alpha \splus \alpha \mid \alpha \stimes \alpha
\end{align*}
where $k \in R$ and $v \in \mathcal{V}$.
\end{definition}
We can evaluate weighted formulas with respect to an interpretation to obtain a value from the semiring.
\begin{definition}[Semantics]
\label{def:wprop}
Given a weighted propositional formula $\alpha$ over a semiring $\mathcal{R} = (R, \srsplus, \srstimes, \srzero, \srone)$ and propositional variables from $\mathcal{V}$ as well as an \emph{interpretation} $\mathcal{I}$, i.e., a subset of $\mathcal{V}$, the semantics $\llbracket \alpha \rrbracket_{\mathcal{R}}(\mathcal{I})$ of $\alpha$ over $\mathcal{R}$ w.r.t.\ $\mathcal{I}$ is defined as follows:
\begin{align*}
    \llbracket k \rrbracket_{\mathcal{R}} (\mathcal{I}) &= k\\
    \llbracket v \rrbracket_{\mathcal{R}} (\mathcal{I}) &= \left\{\begin{array}{cc}
        \srone & v \in \mathcal{I} \\
        \srzero & \text{ otherwise. }
    \end{array} \right. (v \in \mathcal{V})\\
    \llbracket \neg v \rrbracket_{\mathcal{R}} (\mathcal{I}) &= \left\{\begin{array}{cc}
        \srzero &  v \in \mathcal{I} \\
        \srone & \text{ otherwise. }
    \end{array} \right. (v \in \mathcal{V})\\
    \llbracket \alpha_1 \splus \alpha_2 \rrbracket_{\mathcal{R}} (\mathcal{I}) &= \llbracket \alpha_1\rrbracket_{\mathcal{R}} (\mathcal{I}) \srsplus \llbracket\alpha_2 \rrbracket_{\mathcal{R}} (\mathcal{I})\\
    \llbracket \alpha_1 \stimes \alpha_2 \rrbracket_{\mathcal{R}} (\mathcal{I}) &= \llbracket \alpha_1\rrbracket_{\mathcal{R}} (\mathcal{I}) \srstimes \llbracket\alpha_2 \rrbracket_{\mathcal{R}} (\mathcal{I}).
\end{align*}
\end{definition}

We define algebraic measures to combine the qualitative language of answer set programs with the quantitative one of weighted logic.
\begin{definition}[Algebraic Measure]
An \emph{algebraic measure} $\mu = \langle \Pi, \alpha, \mathcal{R}\rangle$ consists of an answer set program $\Pi$, a weighted formula $\alpha$, and a semiring $\mathcal{R}$. Then, the weight of an answer set $\mathcal{I} \in \mathcal{AS}(\Pi)$ under $\mu$ is defined by
\[
\mu(\mathcal{I}) =  \llbracket \alpha \rrbracket_{\mathcal{R}} (\mathcal{I}).
\]
Additionally, the result of an \emph{(atomic) query} for an atom $a$ from $\Pi$ is given by
\[
\mu(a) = \srbplus_{\mathcal{I} \in \mathcal{AS}(\Pi), a \in \mathcal{I}} \mu(\mathcal{I}), 
\]
and the result of the \emph{overall weight query} of $\Pi$ is 
\[
\mu(\Pi) = \srbplus_{\mathcal{I} \in \mathcal{AS}(\Pi)} \mu(\mathcal{I}).
\]
\end{definition}
Intuitively, the idea here is that for an algebraic measure $\mu = \langle \Pi, \alpha, \mathcal{R}\rangle$ the semiring $\mathcal{R}$ determines the mode of quantitative computation, $\Pi$ states the logical background theory that determines which interpretations are solutions and $\alpha$ assigns each answer set $\mathcal{I}$ a weight over the semiring, by performing a calculation over the semiring that depends on the satisfaction of atomic formulas in the interpretation $\mathcal{I}$. 
\subsection{Similarity of Scenes}
Broadly speaking, we can modify a scene in three ways: 
\begin{enumerate}[label={\upshape(\roman*)}, widest=(iii)]
    \item Object Addition,
    \item Object Deletion, or
    \item Object Modification.
\end{enumerate}
Our goal is to assign these actions a cost. Then we can compute how costly it is to obtain one scene from another by performing a sequence of actions and summing up their costs. The higher this cost is the lower is the similarity between the two scenes. 

The effect of (i) and (ii) are clear and do not allow for many suboptions. The only possibility in this direction is to differentiate between the addition/deletion of objects of different complexities, assigning higher costs to more complex objects. This option can be explored more later if necessary. For now, we assume that an addition and deletion have fixed costs $\rem{cost}(Add)$ and $\rem{cost}(Del)$, respectively.

For modification, however, we have different options:
\begin{itemize}
    \item Displacement (e.g., to force an overlap between two objects),%\footnote{Not sure if we need to restrict ourselves to a limited form of displacement here, since we otherwise would need to guess the new position in some form of coordinate system, which sounds costly in ASP.}
    \item Class Variation, and
    \item Property Variation (i.e., removing a property, adding a property, or modifying the value of a property).
\end{itemize}
For class variation it makes sense to add some restrictions, otherwise, we could perform an object modification to achieve a deletion and addition in the same step. For this, we assert that it is only possible to exchange class $C$ by class $C'$ if they share a reasonable superclass $C_s$, i.e., $C \sqsubseteq C_s$ and $C' \sqsubseteq C_s$ must hold for a superclass $C_s$ that is not $\mi{owl\!:\!Thing}$ or something of the sort. Here, we choose a list of reasonable superclasses including $\mi{Vehicle}, \mi{Animal},$ and $\mi{StreetSign}$.

The costs we assign for each of the modifications are as follows:
\begin{itemize}
    \item \textbf{Displacement:} Either the distance between the former and the latter location multiplied by a constant factor or a constant cost $\rem{cost}(Disp)$.
    \item \textbf{Class Variation:} When replacing class $C$ by $C'$ with lowest common superclass $C_s$, the cost is the minimal number of DL-axioms that need to be used to derive that $C \sqsubseteq C_s$ and $C' \sqsubseteq C_s$. For example, consider the following set of axioms:
    \begin{align*}
        \mi{Hedgehog} &\sqsubseteq \mi{DangerousAnimal},\\
        \mi{Tiger} &\sqsubseteq \mi{DangerousAnimal}
    \end{align*}
    Here, for $C = \mi{Hedgehog}, C' = \mi{Tiger},$ and $C_s = \mi{DangerousAnimal}$, we need one DL-Axiom to derive $\mi{Hedgehog} \sqsubseteq \mi{DangerousAnimal}$ and one DL-Axiom to derive $\mi{Tiger} \sqsubseteq \mi{DangerousAnimal}$. Thus, replacing a tiger by a hedgehog would result in a cost of $2$. 
    \item \textbf{Property Variation:} We assign deletion, addition, and modification a constant value $\rem{cost}(PDel),\rem{cost}(PAdd),$ and $\rem{cost}(PMod)$, each. It makes sense to have $\rem{cost}(PDel) = \rem{cost}(PAdd) > \rem{cost}(PMod)$.
\end{itemize}

Given these assumptions on how we measure the similarity, we can proceed with the modelling of its measurement.
\subsection{Measuring Similarity with Algebraic Measures}
Algebraic measures consist of three parts. The program, specifying the logical constraints, the weighted formula specifying how we measure the weight, and the semiring specifying what kind of weight we measure. For the semiring it makes sense to use $\mathcal{R}_{\min, +}$. Then, we can sum up different costs that are incurred by an interpretation and choose the minimum possible cost, if there are different options. 

The logical constraints are mainly given by the MR-CKR. However, in order to specify a weighted formula, we need to be able to use some atomic formulas that tell us which modifications were performed to arrive at the scene in the model. Thus, we ensure that the signature of the program includes the following predicates:
\begin{itemize}
    \item $\rem{addition}(C,I)$, denoting additions of individual $I$ to class $C$;
    \item $\rem{deletion}(C,I)$, denoting deletion of individual $I$ from class $C$;
    \item $\rem{displacement}(I)$ (resp. $\rem{displacement}(I,D)$), denoting displacement of individual $I$ (resp. by distance $D$);
    \item $\rem{classVar}(I, C, C')$, denoting that individual $I$ was in class $C$ in the original scene but is now in class $C'$;
    \item $\rem{propertyVar}(I, T)$, denoting that a property of individual $I$ underwent a modification of type $T$.
\end{itemize}

Using the above mentioned predicates, we can easily specify the weighted formula $\alpha_{cost}$ that measures the cost of transforming the original scene into the modified one, as follows:
\begin{align*}
    &\btimes_{\text{class }c, \text{individual }i} (\rem{addition}(c,i) \stimes \rem{cost}(Add) \splus \neg \rem{addition}(c,i)) \\
    &\stimes \btimes_{\text{class }c, \text{individual }i} (\rem{deletion}(c,i) \stimes \rem{cost}(Del) \splus \neg \rem{deletion}(c,i)) \\
    &\stimes \btimes_{\text{individual }i} (\rem{displacement}(i) \stimes \rem{cost}(Disp) \splus \neg \rem{displacement}(i)) \\
    &\stimes \btimes_{\text{classes }c, c', \text{individual }i} (\rem{classVar}(i, c, c') \stimes \rem{dist}(c,c') \splus \neg \rem{classVar}(i, c, c')) \\
    &\stimes \btimes_{\text{individual }i, t \in \{PDel, PAdd, PMod\}} (\rem{propertyVar}(i, t) \stimes \rem{cost}(t) \splus \neg \rem{propertyVar}(i, t)) 
\end{align*}
One line takes care of the cost for one modification type each. Here, $\rem{dist}(c,c')$ denotes the distances between two classes $c$ and $c'$ as explained above. We can compute these statically for each of the contexts.

\begin{example}
Consider for example the interpretation 
\[
\mathcal{I} = \{\rem{addition}(\mi{RollingContainer},i_1), \rem{deletion}(\mi{Child},i_2), \rem{deletion}(\mi{Child},i_3)\}.
\]
Intuitively, this means that we add the object $i_1$ to the concept $\mi{RollingContainer}$ and remove the objects $i_2$ and $i_3$ from the concept $\mi{Child}$. 

We expect that this interpretation comes with a cost of $\rem{cost}(Add) + 2\cdot \rem{cost}(Del)$ since we add one object to a concept and remove two.

Accordingly, we obtain $\llbracket \alpha_{cost} \rrbracket_{\mathcal{R}_{\min, +}}(\mathcal{I} = \rem{cost}(Add) + 2\cdot \rem{cost}(Del)$. This can be seen as follows. First, observe that the last three rows of the definition of $\alpha_{cost}$ are equal to $\srone$ since $\neg \rem{displacement}(i), \neg \rem{classVar}(i,c,c'), \neg \rem{propertyVar}(i,t)$ hold for all $i,c,c',t$ and thus 
\begin{align*}
& \llbracket \rem{displacement}(i) \stimes \rem{cost}(Disp) \splus \neg \rem{displacement}(i) \rrbracket_{\mathcal{R}_{\min, +}}(\mathcal{I})\\
= & \llbracket \rem{displacement}(i) \stimes \rem{cost}(Disp) \rrbracket_{\mathcal{R}_{\min, +}}(\mathcal{I}) \srsplus\llbracket \neg \rem{displacement}(i) \rrbracket_{\mathcal{R}_{\min, +}}(\mathcal{I})\\
= & \llbracket \rem{displacement}(i) \rrbracket_{\mathcal{R}_{\min, +}}(\mathcal{I}) \srstimes \llbracket \rem{cost}(Disp) \rrbracket_{\mathcal{R}_{\min, +}}(\mathcal{I}) \srsplus\srone \\
= & \srzero \srstimes \llbracket \rem{cost}(Disp) \rrbracket_{\mathcal{R}_{\min, +}}(\mathcal{I}) \srsplus\srone \\
= & \srzero \srsplus \srone = \srone
\end{align*}
The same can be observed for the last two rows. 

On the other hand, by the same reasoning, since the interpretation $\mathcal{I}$ contains $\rem{addition}(\mi{RollingContainer},i_1)$ the first row evaluates to $\rem{cost}(Add)$ and the second row evaluates to $\rem{cost}(Del) \srstimes \rem{cost}(Del)$, resulting in 
\[
    \rem{cost}(Add) \srstimes \rem{cost}(Del) \srstimes \rem{cost}(Del).
\]
Since we use $\mathcal{R}_{\min, +}$ the operation $\srstimes$ is $+$ and we obtain $\rem{cost}(Add) + 2\cdot \rem{cost}(Del)$ as the final cost, as expected.
\end{example}

\subsection{Translation to Weak Constraints}
While algebraic measures are a useful tool, to specify quantitative measures for the answer sets of programs, most solvers for ASP currently do not support optimization of the weight of an algebraic measures. However, the algebraic measure that we use can be translated to \emph{weak constraints}~\cite{DBLP:conf/lpnmr/BuccafurriLR97}.

Recall that intuitively the weighted formula $\alpha_{cost}$ corresponds to the sum of the cost of the modifications that were performed on the scene. E.g.\ for 
\[
(\rem{addition}(c,i) \stimes \rem{cost}(Add) \splus \neg \rem{addition}(c,i))
\]
we either have cost $0$ if $\neg \rem{addition}(c,i)$ holds or cost $\rem{cost}(Add)$ if $\rem{addition}(c,i)$ holds. 

This is exactly the kind of costs that can be modelled and optimized using weak constraints of the form
\[
\text{:$\sim$} a_1, \dots, a_n, \text{not } b_1, \dots, \text{not } b_m. [C,t_1, \dots, t_k],
\]
where $a_i$ and $b_j$ are atom formulas, $C$ is the cost and $t_l$ are terms. This weak constraints means that satisfying $a_1, \dots, a_n$ but not $b_1, \dots, b_m$ incurs a cost of $C$. The terms $t_1, \dots, t_k$ intuitively group different weak constraints. That is, if there are multiple weak constraints with the same terms, then only the one with the highest cost is triggered. 

This means, we can use the weak constraint
\[
\text{:$\sim$} \rem{addition}(C,I). [\rem{cost}(Add),add,C,I]
\]
to add a cost of $\rem{cost}(Add)$ for each individual $i$ and concept $c$ such that $\rem{addition}(c,i)$ holds, i.e., such that we add $i$ to concept $c$.

We can do the same for the other factors of $\alpha_{cost}$ to translate it to ASP with weak constraints.

\section{Implementation Prototype for Scene Generation in Autonomous Driving}
We implemented and tested our approach using an example from Autonomous Driving. Here, we reconstructed and slightly extended the base ontology from \cite{chowdhury2021towards,wickramarachchi2021knowledge}.\footnote{\url{https://github.com/boschresearch/ad_cskg}} It features different scenes each annotated with the objects included in it. The included objects are annotated with information about them, such as their type. Additionally, the ontology includes axioms that add additional knowledge about the relationship between different concepts in the ontology. Overall, the base ontology has more than 3 million axioms, concerning knowledge of 41 concepts, more than 100 scenes, and more than 50 thousand objects. 

In our prototype we restrict ourselves to limited scene variations, i.e., addition and deletion of objects from concepts, and assign them cost $1$ each. However, also the other modifications detailed above could be added without problems. 
\subsection{Overview}
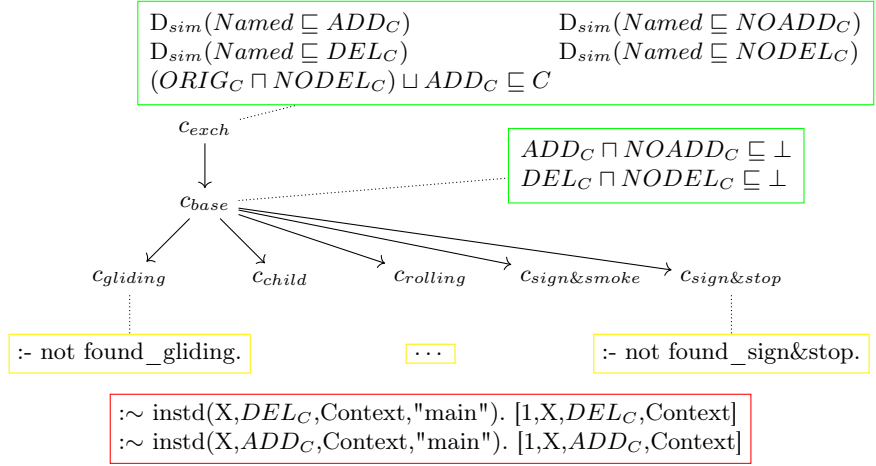
\begin{figure}
    \centering
    \begin{tikzpicture}
        \node (exch) at (-3,0) {$c_{exch}$};
        \node[align=left,rectangle,draw=green] (exchax) at (1,1) {\begin{tabular}{ll}$\default_{sim}(Named \sqsubseteq ADD_{C})$ & $\default_{sim}(Named \sqsubseteq NOADD_{C})$ \\
    $\default_{sim}(Named \sqsubseteq DEL_{C})$ & $\default_{sim}(Named \sqsubseteq NODEL_{C}) $\\
    $(ORIG_{C} \sqcap NODEL_{C}) \sqcup ADD_{C} \sqsubseteq C$\end{tabular}};
        \node (base) at (-3,-1) {$c_{base}$};
        \node[align=left,rectangle,draw=green] (baseax) at (3,-0.5) {\begin{tabular}{l}
    $ADD_{C} \sqcap NOADD_{C} \sqsubseteq \bot$ \\
    $DEL_{C} \sqcap NODEL_{C} \sqsubseteq \bot$\end{tabular}};
        \node (gliding) at (-4,-2) {$c_{gliding}$};
        \node (child) at (-2,-2) {$c_{child}$};
        \node (rolling) at (0,-2) {$c_{rolling}$};
        \node (smoke) at (2,-2) {$c_{sign\&smoke}$};
        \node (stop) at (4,-2) {$c_{sign\&stop}$};
        \node[align=left,rectangle,draw=yellow] (glidingax) at (-4,-3) {:- not found\_gliding.};
        \node[align=left,rectangle,draw=yellow] (otherax) at (0,-3) {$\dots$};
        \node[align=left,rectangle,draw=yellow] (stopax) at (4,-3) {:- not found\_sign\&stop.};
        \node[align=left,rectangle,draw=red] (weakax) at (0,-4) {:$\sim$ instd(X,$DEL_{C}$,Context,"main"). [1,X,$DEL_{C}$,Context]\\
        :$\sim$ instd(X,$ADD_{C}$,Context,"main"). [1,X,$ADD_{C}$,Context]};
        \draw[->] (exch) to (base);
        \draw[->] (base) to (gliding);
        \draw[->] (base) to (child);
        \draw[->] (base) to (rolling);
        \draw[->] (base) to (smoke);
        \draw[->] (base) to (stop);
        \draw[-,densely dotted] (base) to (baseax);
        \draw[-,densely dotted] (exch) to (exchax);
        \draw[-,densely dotted] (gliding) to (glidingax);
        \draw[-,densely dotted] (stop) to (stopax);
    \end{tikzpicture}
    \caption{MR-CKR and additional constraints used in our prototype for autonomous driving. $c_{i}$ denotes the different contexts, arrows between contexts denote higher specificity. Ontology axioms and additional ASP constraints belonging to some context are shown in green and yellow boxes, respectively, and are connected to their context with a dotted line. The red box contains the additional weak constraints that optimize for similarity.}
    \label{fig:prototype-mr-ckr}
\end{figure}

We provide an overall sketch of the MR-CKR and its interplay with ASP constraints in \Cref{fig:prototype-mr-ckr}. We go over the different parts step by step.

As diagnoses for network failure we use the following (using existing classes from the ontology):
\begin{enumerate}
    \item The scene contains an object that is in the class $GlidingOnWheels$. Dangerous due to lack of no examples.
    \item The scene contains an object that is in the class $Child$. Dangerous due to unpredictable behaviour compared to other humans.
    \item The scene contains an object that is in the class $RollingContainer$ but no object in the class $Human$. Dangerous due to unpredictable behaviour of the rolling container.
    \item The scene contains an object that is in the class $Sign$ and an object in the class $Smoke$. Dangerous due to harder recognition of the sign due to smoke.
    \item The scene contains an object that is in the class $Sign$ and an object in the class $StopLineMarking$. Dangerous because the network does not predict stopping at the stop line properly.
\end{enumerate}
This means that we have one context $c_i$ for each diagnosis $i$.

Recall the overall framework from \Cref{fig:framework}. In order to model the possible scene modifications, we have an exchange context $c_{exch}$, that contains for each original (modifiable) concept $C$ in the ontology the following default axioms:
\begin{align*}
    &\default_{sim}(Named \sqsubseteq ADD_{C}) & &\default_{sim}(Named \sqsubseteq NOADD_{C}) \\
    &\default_{sim}(Named \sqsubseteq DEL_{C}) & &\default_{sim}(Named \sqsubseteq NODEL_{C}) \\
    &(ORIG_{C} \sqcap NODEL_{C}) \sqcup ADD_{C} \sqsubseteq C
\end{align*}
Here, 
\begin{itemize}
    \item $Named$ is a concept we define based on the original ontology that contains all modifiable objects, 
    \item $ADD_{C}$ is a concept that represents the individuals that should be added to the concept $C$, 
    \item $NOADD_{C}$ is a concept that represents the individuals that should \emph{not} be added to the concept $C$,
    \item $DEL_{C}$ is a concept that represents the individuals that should be removed from the concept $C$, 
    \item $NODEL_{C}$ is a concept that represents the individuals that should \emph{not} be removed from the concept $C$,
    \item $ORIG_{C}$ is a concept that represents the individuals that were in the concept $C$ in the original ontology.
\end{itemize}
Thus, we assert for each modifiable individual that the should be (not) added/removed to/from the concept $C$. If they were originally in $C$ and are not removed or are added to $C$, then they should be in $C$, as the last axiom asserts. Clearly on its own, this does not make sense, since every modifiable individual will be in every concept. Therefore, we add a base context $c_{base}$ that contains for each (modifiable) concept $C$ in the ontology the following axioms:
\begin{align*}
    &ADD_{C} \sqcap NOADD_{C} \sqsubseteq \bot \\
    &DEL_{C} \sqcap NODEL_{C} \sqsubseteq \bot
\end{align*}
These ensure that we either add (resp.\ remove) or do not add (resp.\ remove) an individual but not both. Then if $c_{exch}$ is less specific than $c_{base}$ with respect to $\succ_{sim}$, we can override the defaults of $c_{exch}$ in $c_{base}$, such that we can satisfy the disjointness requirements in $c_{base}$. 

In order to give the contexts $c_{i}$ for the diagnoses access to the possibility of modification, we then declare each context $c_{i}$ more specific than $c_{base}$ with respect to $\succ_{sim}$. 

What is left, are the additional ASP constraints, that (i) ensure minimal modifications and (ii) ensure the presence of the diagnosis in the given contexts.

For (i), we use the following rules for each modifiable concept $C$:
\begin{minted}[escapeinside=||,mathescape=true]{prolog}
:~ instd(X,|$DEL_{C}$|,Context,"main"). [1,X,|$DEL_{C}$|,Context]
:~ instd(X,|$ADD_{C}$|,Context,"main"). [1,X,|$ADD_{C}$|,Context]
\end{minted}
This ensures that a penalty of $1$ is added every time we add or delete an individual $X$ to $C$. Note that the penalty is applied for every context.

For (ii), we simply add constraints that ensure that the diagnosis is derived. For example, for the third diagnosis in context $C_3$, where we need to derive that there is an object in the concept $RollingContainer$ but none in the concept $Human$, we add the rules
\begin{minted}[escapeinside=||,mathescape=true]{prolog}
found_rolling_no_human_1 :- instd(X, |$RollingContainer$|, |$C_3$|, "main").
:- not found_rolling_no_human_1.
found_rolling_no_human_2 :- instd(X, |$Human$|, |$C_3$|, "main").
:- found_rolling_no_human_2.
\end{minted}

\begin{example}
Assume our input scene contains four objects $i_1, \dots, i_4$ and 
\[
Child(i_2), Child(i_3), Car(i_4)
\]
hold.

Due to the ontology axiom $Child \sqsubseteq Human$, we could derive $Human(i_2)$ and $Human(i_3)$. 

Thus, in context $C_3$, where we need a rolling container but no human, we need to remove $i_2$ and $i_3$ from the $Child$ concept and add an object to the $RollingContainer$ concept. Thus, a potential modification (restricted to context $C_3$) is represented by the interpretation 
\begin{align*}
\mathcal{I} = \{&\rem{instd}(i_1,ADD_{RollingContainer}, C_3, ``main"), \\
    &\rem{instd}(i_2,DEL_{Child}, C_3 ``main"), \\
    &\rem{instd}(i_3,DEL_{Child}, C_3, ``main")\}.
\end{align*}
As discussed in the previous example, it has cost $\rem{cost}(Add) + 2\cdot \rem{cost}(Del)$, which is $3$ since we assign addition and deletion cost $1$. 

Since there is no modification of a lower cost, one of the possible generated scenes for $C_3$ consists of 
\[
RollingContainer(i_1), Car(i_4),
\]
i.e., it contains a rolling container $i_1$, no humans, but a car $i_4$.
\end{example}
\subsection{Practical Implementation}
In order to transfer the idea that we sketched above into a formal encoding of the problem that we can solve with an ASP solver such as clingo~\cite{gebser2014clingo}, we proceed in the following steps:

\begin{enumerate}
    \item Build an MR-CKR encoding
    \item Translate the MR-CKR encoding to ASP
    \item Add strong constraints to ensure danger
    \item Add weak constraints to ensure similarity
\end{enumerate}
We implemented all these steps and made them available online.\footnote{\url{https://github.com/raki123/MR-CKR}} In more detail, we tackle them as follows.

\paragraph{Building an MR-CKR encoding.} Here, we read the base ontology and gather (i) axioms that generally hold and (ii) knowledge regarding some particular scene. Then, we build the MR-CKR as sketched in \Cref{fig:prototype-mr-ckr}. There are two details to note here. First, we additionally add the general axioms from the base ontology that we previously gathered to $c_{base}$. This ensures that the generated scenes are realistic. Second, we do not add defaults to add/delete individuals to concepts for every concept but only a subset of relevant ones. This helps us by reducing the size of the problem encoding and by improving inference performance.

\paragraph{Translating the MR-CKR encoding to ASP.} The CKRew software\footnote{\url{https://github.com/dkmfbk/ckrew}} is an existing tool from previous work~\cite{DBLP:journals/tplp/BozzatoEK21} that performs the desired translation of MR-CKRs to ASP. The original translation is capable of handling highly complex relations between contexts and supports arbitrary defaults and flexible ontological background knowledge. However, this comes at the cost of an encoding in ASP that is not suitable for our purposes, using large scene graphs with many contexts and concepts. 

To circumvent this, we specialized the encoding to our setting. Namely, for a given concept $C$, the (defeasible) axioms 
\begin{align*}
    &\default_{sim}(Named \sqsubseteq ADD_{C}) & &\default_{sim}(Named \sqsubseteq NOADD_{C}) \\
    &ADD_{C} \sqcap NOADD_{C} \sqsubseteq \bot
\end{align*}
tell us that we guess \emph{either} the addition \emph{or} the non-addition of any named individual $X$ to $C$, as long as there is no other reason in the ontology that prevents both. Since our base ontology is consistent, there can never be a reason in our ontology that prevents the non-addition. Therefore, the either-or really holds in our setting. 

This allows us to use the following rules to encode the (defeasible) axioms above:
\begin{minted}[escapeinside=||,mathescape=true]{prolog}
instd(X,|$ADD_{C}$|,Con,"main") :- instd(X,"Named",Con,"main"), 
                                not instd(X,|$NOADD_{C}$|,Con,"main").
instd(X,|$NOADD_{C}$|,Con,"main") :- instd(X,"Named",Con,"main"), 
                                not instd(X,|$ADD_{C}$|,Con,"main").
\end{minted}
We do the same for deletion and non-deletion. 

This specialized translation for our setting leads to a significant performance improvement. While the original encoding only allows us to generate new scenes using tiny starting scenes, the improved strategy allows inference of the real world scenes from the ontology within seconds.

\paragraph{Adding ASP constraints} The addition of the strong and weak constraints is surprisingly simple. We can refer to the derived knowledge by making use of the vocabulary that CKRew uses for translation. Thus, we can provide all additional constraints in a separate program file and solve the combination of the ASP encoding of the MR-CKR and the additional constraints.

\subsection{Scalability}
We briefly investigate how large the instances that we can solve can become, while maintaining a low runtime. Here, we consider on the one hand the original translation of MR-CKR to ASP, denoted \textsc{General}, and on the other hand the specialized translation that makes use of the restricted use of defaults, denoted \textsc{Specialized}. The aim here is to show that is not only helpful but even necessary to use \textsc{Specialized} over \textsc{General}, when solving real world problems.

Secondly, we investigate the dependence of the runtime on (i) the number of objects in the scene and (ii) the number of contexts. We vary (i) between 1 and 20 and (ii) between 1 and 5 on a randomly chosen example scene from the ontology.

We only measure the solving time, since building the MR-CKR and translating the MR-CKR to an ASP encoding only consumed insignificant time (less than 1 second) regardless of the translation and number of objects/contexts. 

For the solving phase, we use ``clingo''~\cite{gebser2014clingo} on the non-ground program with input option ``-t 3'' to specify that three threads in parallel should be used to solve the problem. We apply a time limit of 120 seconds and assign runs that do not finish during this time a runtime of 120 seconds.

\begin{figure}[t]
    \centering
    \includegraphics[width=1.0\textwidth]{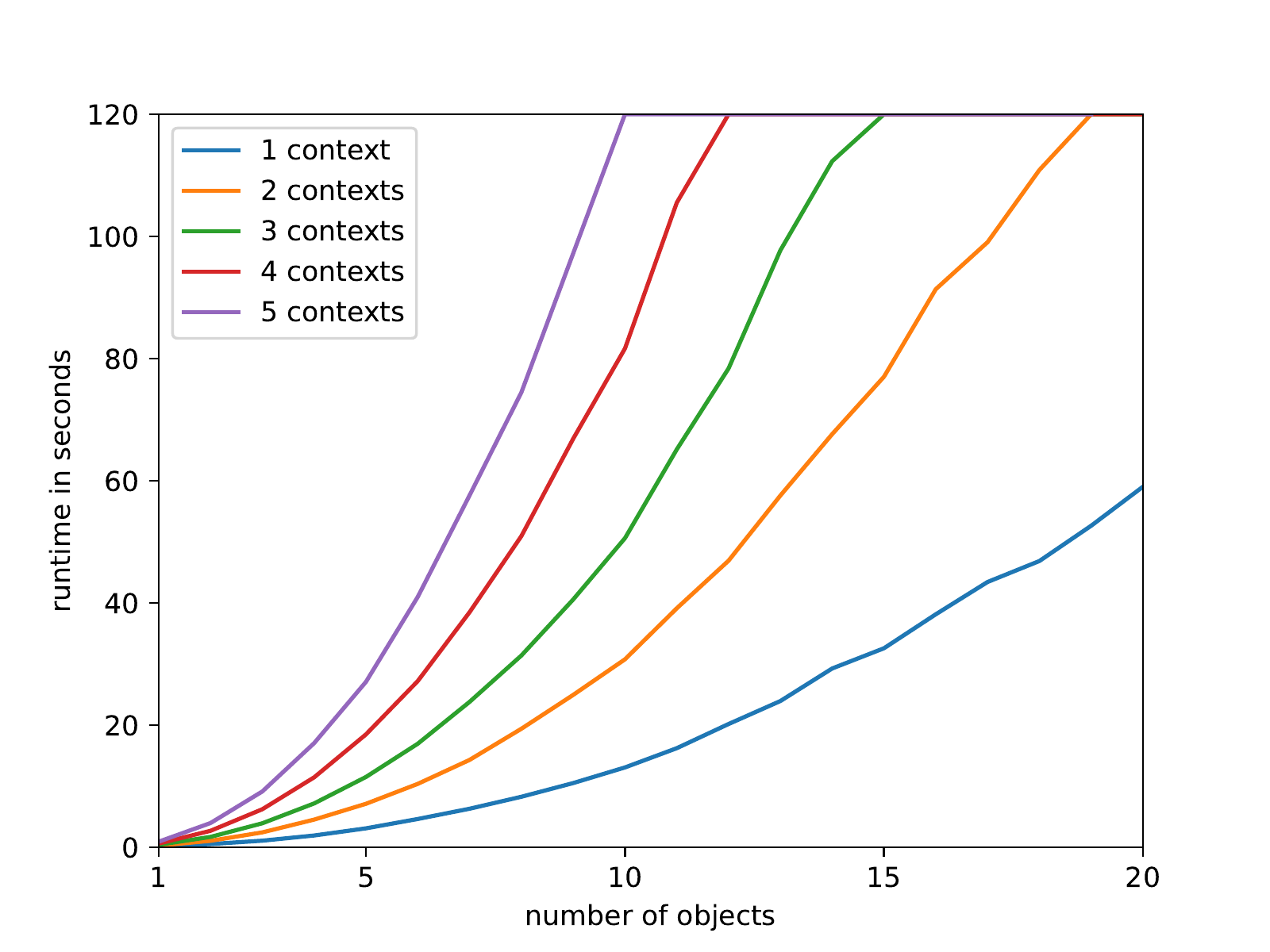}
    \caption{Solving time after using the \textsc{General} translation of MR-CKR to ASP.}
    \label{fig:naive}
\end{figure}
\begin{figure}[t]
    \centering
    \includegraphics[width=1.0\textwidth]{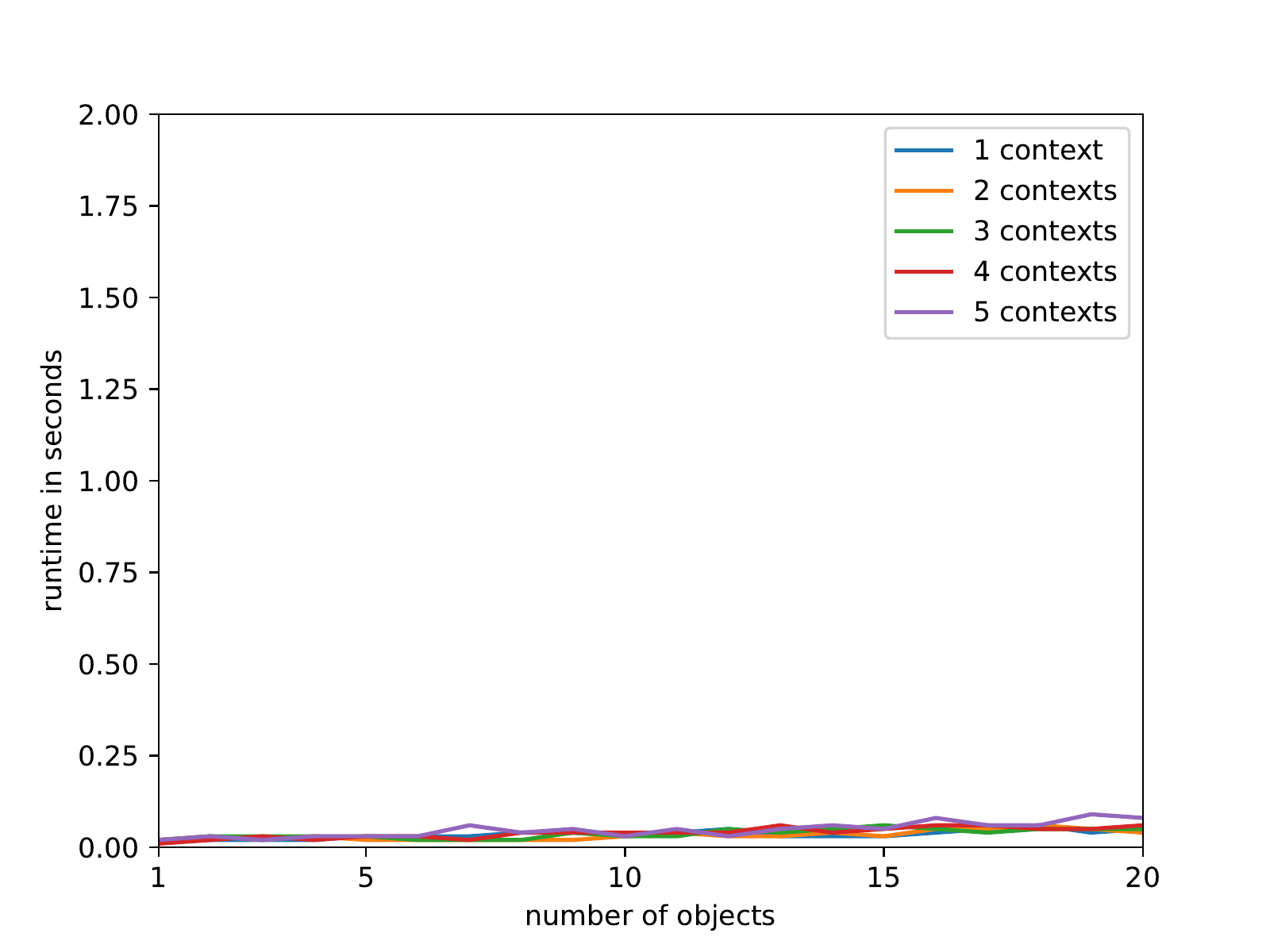}
    \caption{Solving time after using the \textsc{Specialized} translation of MR-CKR to ASP.}
    \label{fig:specialized}
\end{figure}
The results of our investigation are given in \Cref{fig:naive,fig:specialized}. We see that even if only one context is used, the runtime after \textsc{General} grows quickly. While it still remains in a feasible range, when using one context and up to 20 objects in the scene, the same cannot be said when more contexts are used. For five contexts, solving already becomes slow when ten or more objects are included in the scene. Additionally, the original scene has many more objects (more than 300), thus, this translation can only be employed to restricted examples, even if there is only one context.

On the other hand, for \textsc{Specialized} we see that the solving time is consistently far below one second, even when using all five contexts and 20 objects in the scene. Note here the different limits of the Y-axis, which we adapted to make the runtimes visible. Even when using all objects that were originally included in the scene (more than 300) the solving time remains at around 0.67 seconds.

We see that while the original translation \textsc{General} is able to handle a broader range of MR-CKRs, it pays off to use the specialized translation \textsc{Specialized} in our setting. With \textsc{Specialized} we can generate new scenes in subsecond times, even if the full scene (i.e.\ all its objects) and all contexts are used. This suggests that with \textsc{Specialized} we can also generate new inputs for more complex semantic conditions and base ontologies than the ones provided in our prototype, giving us interesting opportunities to extend our work in the future.

\section{Conclusion}
We introduced a new framework to generate new interesting inputs for neural models based on existing ones, in particular the setting of scene generation for AV scene data. Notably, our framework does so based on symbolic reasoning methods: this allows us, on the one hand, to incorporate real world knowledge (in the form of contextual knowledge) that ensures that the generated inputs are \emph{realistic}, and, on the other hand, to formulate a semantic criterion that should be satisfied by the new input.

We saw that all components that we incorporated in our framework add their respective benefits:
\begin{itemize}
    \item \textbf{MR-CKR} allows us (i) to incorporate ontological knowledge easily and (ii) to perform different modifications in different contexts.
    \item \textbf{Algebraic Measures} allow us to easily specify a cost value to optimize.
    \item \textbf{ASP}, as a declarative programming language to translate to, allows us to perform reasoning/scene generation efficiently using standard solvers.
\end{itemize}
While we only considered a small example in our prototype, it successfully generates new scene descriptions. Furthermore, as it can be easily generalized to the generation of different types of scenes, it provides a proof of concept of our approach.

In future work, it will be interesting to extend this example with more complicated semantic descriptions of interesting scenes gathered by inspecting poor performing inputs for a prediction task with a neural model: in particular, it would be interesting to use more complex contextual structures to represent different variations of the scenes, but also use inputs performances to give a quantification of the more interesting cases to be generated. 
Another open challenge is to use the symbolic description of the new scene to generate images that can be fed to the neural model and assess how much training with these new examples improves the network performance.

\section*{Acknowledgements}
  This work was partially supported by the European Commission funded
	projects ``Humane AI: Toward AI Systems That Augment and Empower Humans by
	Understanding Us, our Society and the World Around Us'' (grant \#820437) and
	``AI4EU: A European AI on Demand Platform and Ecosystem'' (grant \#825619),
	and the Austrian Science Fund (FWF) project W1255-N23. The support is
	gratefully acknowledged.
 \newpage

\nocite{BozzatoES:18}
\bibliographystyle{splncs04}
\bibliography{bibliography}

%---------------------------------------------------
\appendix
%\section{Rule tables}

\end{document}